\documentclass{article}

\PassOptionsToPackage{numbers, compress}{natbib}

\usepackage[main, final]{neurips_2026}

\usepackage[utf8]{inputenc} 
\usepackage[T1]{fontenc}    
\usepackage{hyperref}       
\usepackage{url}            
\usepackage{booktabs}       
\usepackage{amsfonts}       
\usepackage{nicefrac}       
\usepackage{microtype}      
\usepackage{xcolor}         

\usepackage{graphicx}

\usepackage{eso-pic}
\usepackage{xparse}

\newlength{\badgewidth}
\newlength{\badgegap}
\newcommand{\badgeList}{}

\NewDocumentCommand{\addTopRightBadge}{O{} m}{%
\gappto{\badgeList}{\href{#1}{\includegraphics[width=\badgewidth]{#2}}\hspace{\badgegap}}%
}

\newcommand{\placeTopRightBadges}{%
\AddToShipoutPictureBG*{%
\put(\LenToUnit{\paperwidth - 8cm - \badgewidth},\LenToUnit{\paperheight - 2cm}){%
\makebox[0pt][r]{\badgeList}%
}%
}%
}

\addTopRightBadge{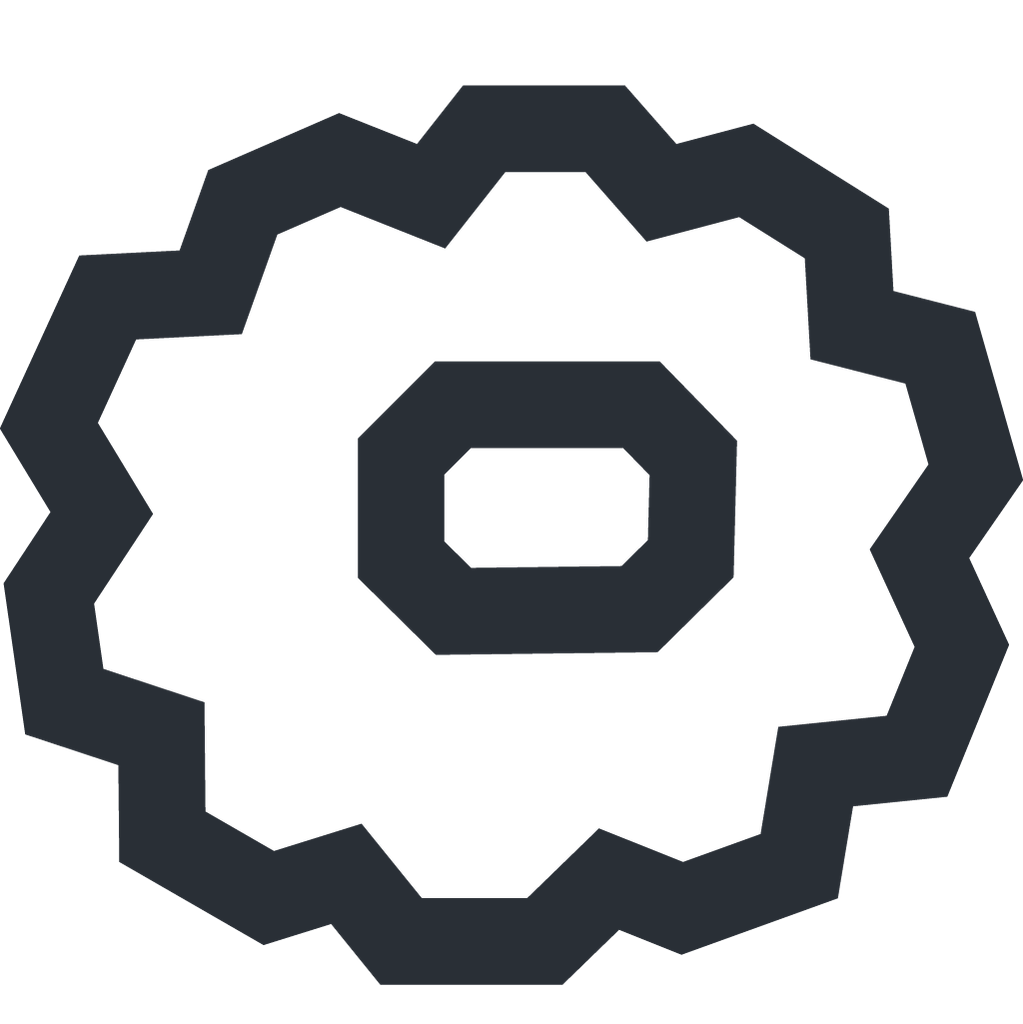}

\placeTopRightBadges

\title{Flower Hub: A Reproducible Benchmarking Platform for Federated Learning in Simulation and Deployment}

\author{%
  Yan Gao\textsuperscript{1,2,}\thanks{Equal contribution}, Mohammad Naseri\textsuperscript{1,*}, Javier Fernandez-Marques\textsuperscript{1,2}, Dimitris Stripelis\textsuperscript{1}, \\
  \textbf{Lorenzo Sani\textsuperscript{1,2}, Davide Eynard\textsuperscript{3}, Fan Zhang\textsuperscript{2}, Hong Jia\textsuperscript{4}, Ting Dang\textsuperscript{5}, D. B. Emerson\textsuperscript{6},} \\
  \textbf{Fatemeh Tavakoli\textsuperscript{6}, Ole Werger\textsuperscript{7}, Lars Wulfert\textsuperscript{7}, Petros Demetrakopoulos\textsuperscript{8},} \\
  \textbf{Sofia Tsekeridou\textsuperscript{8}, InSeo Song\textsuperscript{9}, KangYoon Lee\textsuperscript{9}, Honghao Li\textsuperscript{10}, Lingjuan Lyu\textsuperscript{11},} \\
  \textbf{John P Dickerson\textsuperscript{3}, Daniel Janes Beutel\textsuperscript{1}, Nicholas D. Lane\textsuperscript{1,2}} \\
   \\
  \small \textsuperscript{1}Flower Labs, \textsuperscript{2}University of Cambridge, \textsuperscript{3}Mozilla.ai, \textsuperscript{4}University of Auckland, \\
  \small \textsuperscript{5}University of Melbourne, \textsuperscript{6}Vector Institute, \textsuperscript{7}Fraunhofer IMS, \textsuperscript{8}NetCompany, \\
  \small \textsuperscript{9}Gachon University, \textsuperscript{10}Owkin, \textsuperscript{11}Sony AI \\
}

\begin{document}

\maketitle

\begin{abstract}
Federated learning (FL) has emerged as a key approach for training models across decentralized data, yet benchmarking in FL remains difficult to reproduce, compare, and extend. Existing evaluations are often tied to custom infrastructure, released as incomplete research code, and conducted primarily in simulation, which limits portability and practical relevance. We present Flower Hub, a platform for publishing, discovering, and executing decentralized and federated applications. We show how it enables reproducible benchmarking by packaging benchmarks as executable, versioned applications with standardized metadata, pinned dependencies, and explicit evaluation workflows. We instantiate this approach with a multi-domain benchmark suite spanning cross-silo and cross-device settings, and including tasks in medical imaging, financial tabular learning, legal instruction tuning, phishing URL detection, and audio tagging. We further demonstrate that the same benchmarking application can run across both simulation and deployment runtimes without changing the application code, enabling unified evaluation across varying learning environments. Beyond model quality, our benchmark design supports system-aware reporting, including runtime and communication metrics. This work advances benchmarking in FL settings from ad hoc code artifacts towards portable, executable, and reusable benchmark applications.
\end{abstract}

\section{Introduction}

Many modern machine learning applications rely on distributed data that cannot be centralized due to privacy, regulatory, or bandwidth constraints~\citep{gao2025flowertune,saab2024capabilities,yang2023fingpt,zhang2021survey}. This occurs in domains such as healthcare, edge devices, finance, and legal services~\citep{imteaj2021survey,saab2024capabilities,yang2023fingpt,zhang2023fedlegal}. Federated learning (FL)~\citep{li2020federated,mammen2021federated,mcmahan2017communication,zhang2021survey} has emerged as a key paradigm for training models across decentralized datasets without requiring raw data to be moved to a central location. In recent years, FL research has grown rapidly, with new algorithms, optimization methods, privacy mechanisms, and system architectures proposed at an increasing pace~\citep{ji2024emerging,zhang2021survey}. As the number of FL approaches continues to expand, fair comparison and validation under unified benchmark settings have become increasingly important.

Despite this progress, benchmarking FL systems is inherently challenging. Unlike centralized ML experiments, FL studies typically require infrastructure for collaborative training across multiple parties. As a result, researchers must often manage not only the algorithms being evaluated but also the underlying distributed systems logic. Many existing FL benchmarks are isolated projects that tightly couple infrastructure with application code~\citep{caldas2018leaf,lai2022fedscale,liang2020flbench,ogier2022flamby,ye2024fedllm}. Consequently, sharing a benchmark with another team often requires sharing the entire infrastructure stack rather than only the model, dataset, and algorithmic components. This coupling makes benchmarks difficult to reuse, extend, and reproduce. Reproducing FL results across teams, therefore, demands substantial engineering effort and deep system expertise, which undermines reproducibility, a central requirement of any benchmark.\looseness-1

Another limitation is that FL evaluation is still dominated by simulation~\citep{chai2020fedeval,karargyris2023federated}. While simulation is useful for preliminary algorithmic assessment, transitioning from simulation to real-world federated deployment often requires substantial code changes and major engineering overhead. More importantly, FL lacks a standardized benchmark packaging format. Existing benchmarks are commonly released as research repositories, custom scripts, and partially documented pipelines, with no unified way to package training code, evaluation protocols, distributed configurations, and runtime environments. As a result, each new benchmark often becomes another isolated project, thereby limiting the scalability and long-term usability of FL benchmarking efforts. Finally, many FL papers continue to evaluate their methods on centralized datasets such as CIFAR-10 and MNIST~\citep{li2021model,li2022federated,rehman2023dawa,zhao2018federated}. Although these datasets are useful for initial experimentation, they do not adequately capture the properties of real-world FL deployments, such as statistical heterogeneity, decentralized data ownership, domain-specific constraints, and practical system limitations.

\begin{figure}[t]
\centering
    \includegraphics[width=0.99\linewidth]{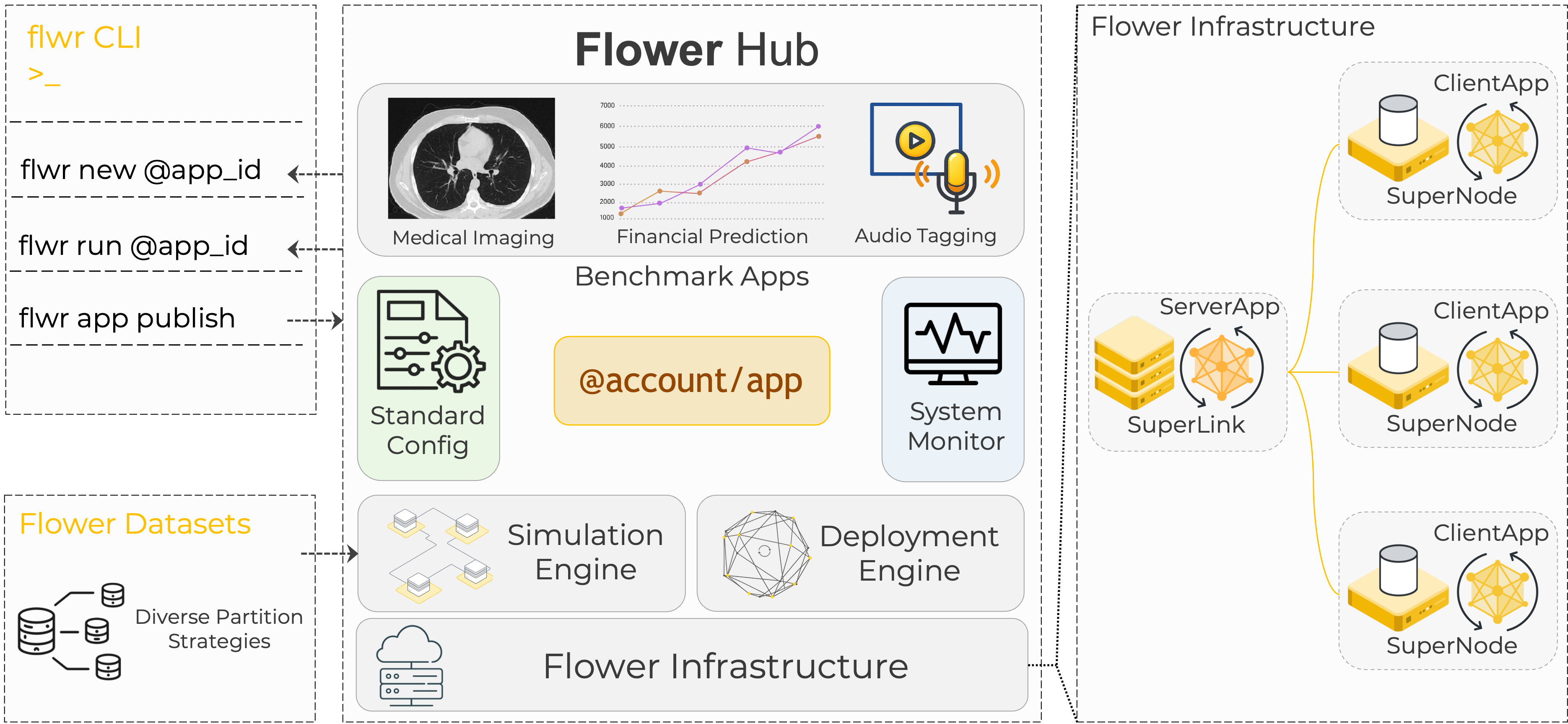}
\caption{\small \textbf{Overview of Flower Hub.} Flower Hub is an open-source benchmarking platform for federated and decentralized learning that decouples application logic from infrastructure, enabling lightweight, portable and shareable benchmark applications. This design allows researchers to focus on algorithmic development and evaluation. Apps can be executed smoothly across both simulation and real-world deployment settings without code modifications, thereby accelerating the transition from development to deployment. Each published benchmarking app follows a standardized package structure, enhancing reproducibility and facilitating sharing, reuse, and extensibility. An integrated system monitor provides real-time tracking of key performance metrics. Together, these features position Flower Hub as a foundation for a collaborative federated benchmarking ecosystem.\looseness-1}
\label{fig:pipeline}
\vspace{-1mm}
\end{figure}

To address these challenges, we propose Flower Hub, an executable and reproducible benchmarking platform for federated and decentralized AI. Flower Hub decouples infrastructure logic from application logic by leveraging the underlying Flower FL infrastructure~\cite{beutel2020flower}. This design enables researchers to focus on the essential components of their work, such as model training, algorithm design, and evaluation, while making benchmarks lightweight to share because no custom infrastructure code is required. Flower Hub provides portability by command across simulation and deployment settings without code changes. It also introduces a standardized benchmark package that specifies training, evaluation, system configuration, and versioning, thereby improving reproducibility and reusability across benchmarks. In addition, Flower Hub provides users with tools to measure system-level metrics, such as communication cost and latency. Together, these features establish Flower Hub as an open-source community ecosystem in which researchers can publish executable federated benchmarks that others can run, evaluate, and build upon. In this sense, Flower Hub represents an ``\textit{App Store moment}” for federated and decentralized benchmarking.

To demonstrate the utility of Flower Hub, we introduce five realistic FL benchmarks spanning five domains: healthcare, finance, security, automotive, and legal services. Our selected benchmarks also cover various data modalities, including text, tabular data, imaging, and audio. Using these benchmarks, we evaluate six widely used FL optimization algorithms in both simulation and deployment settings. The resulting analysis provides insights into the feasibility and effectiveness of deploying FL algorithms in realistic settings. More broadly, these benchmarks can help accelerate the development of inclusive, privacy-preserving, and domain-specialized models for real-world applications.\looseness-1

\section{Flower Hub}

Flower Hub\footnote{\scriptsize\url{https://flower.ai/apps}\label{fn:apps_page}} is an open-source benchmarking platform for federated learning. It supports the discovery, distribution, and execution of benchmarks across both simulation and deployment environments. In Flower Hub, each benchmark is represented as an executable application, allowing users to publish, download, and run benchmarks through a unified interface. This design shifts FL benchmarking from isolated, one-off research efforts toward a collaborative ecosystem that promotes reproducibility and reusability (Figure~\ref{fig:pipeline}). Further details on using Flower Hub are provided in Appendix~\ref{appdix:hub_howto}.\looseness-1

\subsection{Decoupling Application Logic from Infrastructure}

Flower Hub is built on top of the Flower infrastructure, in which a long-running \textit{SuperLink} process represents the central server, while multiple long-running \textit{SuperNode} processes represent federated clients. Further details of the Flower infrastructure are provided in the Appendix~\ref{appdix:flwr_infra}. By relying on this shared infrastructure layer, Flower Hub allows researchers to focus exclusively on application logic.
Specifically, the \textit{ServerApp} defines the overall FL workflow, including aggregation strategies and, when applicable, centralized evaluation. The \textit{ClientApp} defines local training, local evaluation, and the corresponding data pipeline. Benchmark apps on Flower Hub therefore need only to include a \textit{ServerApp} and a \textit{ClientApp}. This design makes apps lightweight to distribute and allows researchers to focus on algorithmic and experimental design without implementing or maintaining infrastructure code.\looseness-1

\subsection{One-Command Portability Across Simulation and Deployment}

Benchmark apps on Flower Hub can be executed in both simulation and deployment settings without code modifications, using a unified command: \textit{flwr run}. In simulation mode, the Flower Simulation Engine runs the full FL workflow locally by launching a \textit{SuperLink} and multiple \textit{SuperNodes}, enabling efficient experimentation even on a single machine.
In deployment mode, the same workflow is executed across distributed environments, where independently launched \textit{SuperNodes} connect to a \textit{SuperLink}. The app is downloaded and executed without changes, significantly reducing the gap between development and real-world deployment.

To support both modes, each benchmark provides dedicated data-loading logic. Simulation uses the Flower Datasets library (Appendix~\ref{appdix:flwr_datasets}) for flexible partitioning, while deployment loads local data from disk. Consistent partitioning and preprocessing ensure alignment between simulation and deployment settings.

\subsection{Standardized Benchmark Package}

Every benchmark app on Flower Hub follows a standardized package structure with a specific schema, explicit configurations, pinned dependencies, and machine-readable metadata. The configuration file includes both task-related and system-related settings. Task-related configurations specify the dataset, model architecture, training hyperparameters, and aggregation strategy. System configurations specify the partitioning scheme, client participation rate, number of training rounds, and client resource requirements for simulation. Defining these components in a single configuration file improves reproducibility and makes benchmarks easier to share, reuse, and extend (see Appendix~\ref{appdix:config_example} for an example).\looseness-1

Flower Hub also provides a flexible version-control mechanism for benchmark apps. Publishers can release updated versions of their benchmarks, while users can download and run specific versions from the Hub. In addition, Flower Hub can automatically identify a compatible app version based on the user’s local environment, further improving usability and reproducibility across different systems.

\subsection{System and ML Performance Monitoring}

Monitoring is essential in FL, both to ensure reliable execution and to understand algorithmic behavior. Flower Hub provides a dedicated monitoring framework for system-level metrics in real time. These metrics include end-to-end round latency, client training time, server aggregation time, communication cost, and client-side CPU and GPU memory utilization. Such measurements enable researchers to assess system health during benchmarking, diagnose performance bottlenecks, and better understand the practical trade-offs of different FL configurations.

At the same time, benchmark apps can report task-specific machine learning metrics through their \textit{ServerApp} and \textit{ClientApp} logic. These metrics can include local and centralized loss or accuracy, convergence statistics, class-wise performance, and other algorithm-specific quantities such as gradient- or pseudo-gradient-related noise measures. This separation preserves Flower Hub's role as a general benchmarking and execution platform while allowing each benchmark to expose the ML observables most relevant to its task and method. Together, system-level and ML-level measurements provide a more complete basis for evaluating federated learning benchmarks.

\section{Benchmark Development}

To demonstrate the utility of Flower Hub, we introduce five realistic FL benchmark tasks: medical image segmentation, financial fraud detection, legal instruction tuning, phishing URL detection, and on-device audio tagging. These tasks involve sensitive, distributed data across institutions or user devices, making FL a natural framework for collaborative training.
Collectively, these benchmarks cover both cross-silo and cross-device FL scenarios, spanning image, tabular, audio, and text modalities. For each task, we establish a complete training and evaluation pipeline and use it to benchmark six widely adopted FL optimization algorithms under standardized experimental settings.

\begin{table}[t]
    \caption{\small Summary of the federated learning benchmark datasets, including their application domain, data modality, number of samples, number of clients, and FL setting.}
    \label{tab:datasets_sum}
    \centering
    \resizebox{0.8\columnwidth}{!}{
    \begin{tabular}{lccccc}
         \toprule
         Datasets & Domain & Data modality & Total \# samples & \# clients & FL type \\
         \hline
         fed-brats\textsuperscript{\ref{fn:med}} & Medical & Image & 1.6 K & 5 & Cross-silo \\
         fed-fraud-paysim-banks\textsuperscript{\ref{fn:fin}} & Finance & Tabular & 6.4 M & 5 & Cross-silo \\
         fed-legal\textsuperscript{\ref{fn:law}} & Law & Text & 83.6 K & 5 & 
         Cross-silo \\
         fed-phishing-urls\textsuperscript{\ref{fn:url}} & Security & URL & 1.1 M & 100 & Cross-device \\
         fed-urbansound8k\textsuperscript{\ref{fn:audio}} & Sensing & Audio & 8.7 K & 50 & Cross-device \\
         \bottomrule
    \end{tabular}
    }
\end{table}

\subsection{FL Dataset Construction}

We carefully select five realistic FL datasets designed to reflect practical deployment environments. The details of each benchmark are described below, with a summary of the proposed datasets presented in Table~\ref{tab:datasets_sum}. A summary of the training and evaluation pipelines for the benchmark tasks is provided in Table~\ref{tab:train_eval}. The distribution of samples across clients is illustrated in Figure~\ref{fig:data_samples}, while client-level label distributions are provided in the Appendix~\ref{appdix:label_dist}.

\textbf{Medical image segmentation}\footnote{\scriptsize\url{https://huggingface.co/datasets/flwrlabs/fed-brats}\label{fn:med}}. We use the BraTS-GLI dataset~\citep{menze2014multimodal}, which comprises multimodal glioma MRI scans with four input sequences: T1-native, T1-contrast enhanced, T2-weighted, and T2-FLAIR, together with corresponding tumor segmentation masks. To emulate a realistic federated setting, the data are partitioned according to acquisition site, with each site representing an individual federated client. Within each site, samples are randomly divided into training and held-out evaluation subsets using an approximate 80:20 split, while ensuring at least one training sample for sites containing multiple cases. This site-wise partitioning preserves institutional heterogeneity and avoids artificial mixing of data across centers, thereby better reflecting the non-IID conditions commonly encountered in multi-institutional medical imaging federated learning.

\textbf{Financial fraud detection}\footnote{\scriptsize\url{https://huggingface.co/datasets/flwrlabs/fed-fraud-paysim-banks}\label{fn:fin}}. We use a PaySim-style synthetic fraud detection dataset~\citep{synthetic_financial_fraud_detection} consisting of transaction-level records with binary fraud labels. To emulate a realistic federated banking environment, transactions are partitioned across five simulated banks using account-level assignment based on the originating account identifier. This ensures that all transactions associated with a given account remain within a single client, preventing account-level leakage across institutions. Client heterogeneity is introduced through predefined quotas that vary bank size, fraud prevalence, and active non-fraud account composition. Within each bank, a stratified train--test split allocates approximately 10\% of fraudulent and non-fraudulent transactions to the held-out test set. The resulting benchmark captures key characteristics of cross-institutional financial FL, including client imbalance, heterogeneous fraud distributions, and institution-specific account ownership.

\textbf{Legal instruction tuning}\footnote{\scriptsize\url{https://huggingface.co/datasets/flwrlabs/fed-legal}\label{fn:law}}. We construct a federated legal supervised fine-tuning benchmark from five legal NLP sources: LexGLUE LEDGAR, LexGLUE CaseHOLD, LexGLUE Unfair Terms of Service, LexGLUE SCOTUS~\citep{chalkidis2022lexglue}, and merged LegalBench~\citep{guha2023legalbench} contract natural language inference tasks. To emulate a realistic federated legal setting, each source task is assigned to a separate client silo, such that clients differ not only in data samples but also in task formulation, label space, and legal subdomain. This creates a strongly non-IID partition reflecting institutional specialization across legal organizations. Each client's dataset is converted into a chat-style supervised fine-tuning format and partitioned locally into training, validation, and test subsets using a deterministic 90:5:5 split. Global validation and test sets are then formed by concatenating held-out subsets from all clients, preserving client-level heterogeneity while enabling centralized evaluation.

\begin{figure}[t]
\centering
    \includegraphics[width=\linewidth]{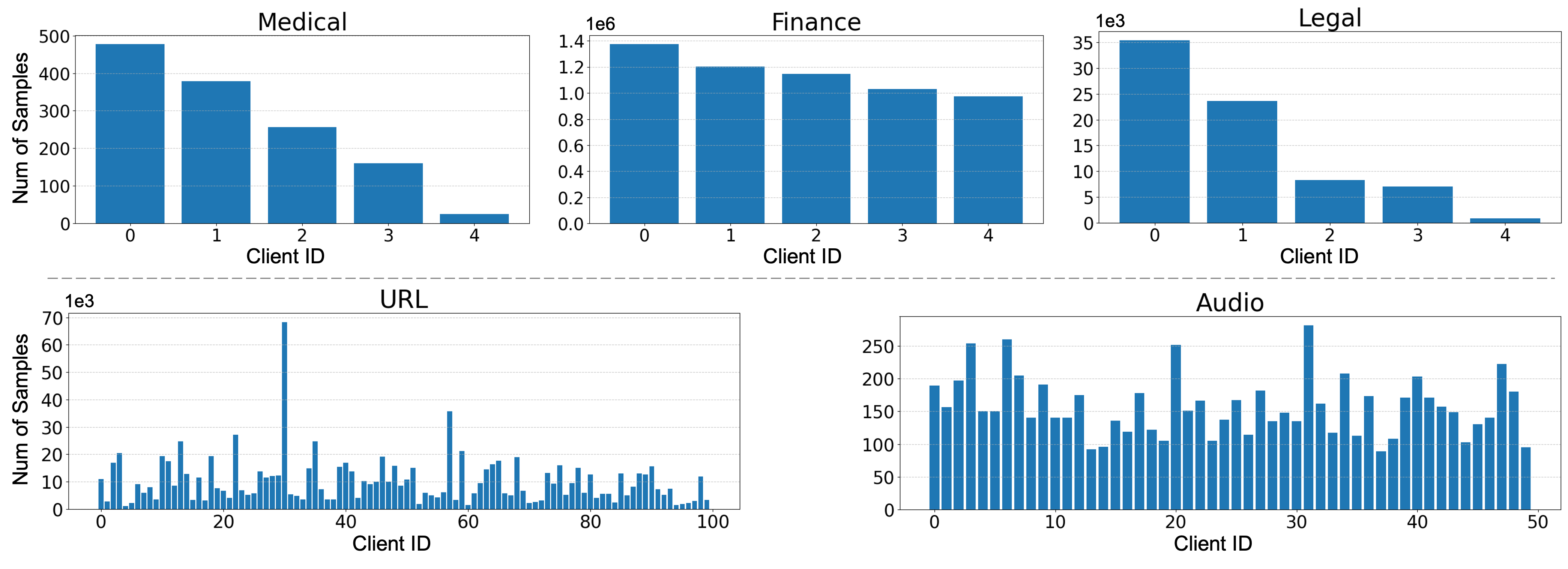}
\caption{\small Distribution of sample counts across clients for the five proposed benchmark datasets.}
\label{fig:data_samples}
\vspace{-1mm}
\end{figure}

\textbf{Phishing URL detection}\footnote{\scriptsize\url{https://huggingface.co/datasets/flwrlabs/fed-phishing-urls}\label{fn:url}}. Phishing detection in real-world cybersecurity systems is inherently decentralized, with URL data distributed across users, organizations, and network endpoints, each exhibiting distinct traffic patterns and threat exposure. This leads to strong heterogeneity in both feature distributions and phishing prevalence across clients, making it a natural and challenging setting for FL.
To reflect these characteristics, we construct a federated phishing URL detection benchmark by merging two public Hugging Face datasets containing URL strings with binary phishing labels~\citep{alvarado_phishing_dataset,kmack_phishing_urls}. Prior to partitioning, URLs are canonicalized and deduplicated. We then allocate samples across 100 simulated clients to mimic decentralized data ownership. Client heterogeneity is introduced through imbalanced client sizes, varying phishing rates sampled from a Beta distribution, and feature-level skew based on URL properties such as suspicious domains, URL shorteners, IP-based hosts, and complex query patterns. Each client is subsequently split into local training and test sets using a 90:10 partition.
This design captures both label imbalance and feature heterogeneity observed in real-world phishing detection, providing a realistic and challenging benchmark for FL.

\textbf{On-device audio tagging}\footnote{\scriptsize\url{https://huggingface.co/datasets/flwrlabs/fed-urbansound8k}\label{fn:audio}}. We construct a federated version of the UrbanSound8K audio classification dataset~\citep{salamon2014dataset}. To emulate a realistic federated acoustic sensing scenario, samples are distributed across 50 clients while preserving fsID groups, such that all clips associated with the same source identifier remain within a single client. This prevents leakage of related recordings across clients and reflects settings in which data originate from distinct devices or environments. Client heterogeneity is introduced through imbalanced log-normal client sizes and client-specific label preferences sampled from a Dirichlet distribution, with assignment performed at the fsID-group level. Each client is subsequently partitioned into local training and test subsets using an approximate 90:10 split, with label-stratified allocation where feasible. The resulting benchmark preserves source-level grouping, client imbalance, and label-distribution heterogeneity, providing a realistic testbed for cross-device federated audio classification.

\begin{table}[t]
    \caption{\small Summary of the training and evaluation pipelines for the benchmark tasks. “CLS” denotes classification, and “Trainable” refers to the number of trainable model parameters.}
    \label{tab:train_eval}
    \centering
    \resizebox{0.8\columnwidth}{!}{
    \begin{tabular}{lccccc}
         \toprule
         Tasks & Task type & Model & Model size & Trainable & Evaluation metrics \\
         \hline
         Medical imaging & Segmentation  & 3D U-Net & 1.77 M & 1.77 M & Dice score \\
         Financial fraud detection & Binary CLS & MLP & 10.75 K & 10.75 K & PR-AUC  \\
         Legal instruction tuning & Instruction tuning & LLM & 3.08 B & 7.67 M & F1 score  \\
         Phishing URL detection & Binary CLS & 1D CNN & 0.57 M & 0.57 M & ROC-AUC \\
         On-device audio tagging & CLS & CNN & 24.40 K &24.40 K & Accuracy \\
          
         \bottomrule
    \end{tabular}
    }
\end{table}

\subsection{Training and Evaluation Pipeline}

To validate the proposed federated datasets, we establish lightweight training pipelines for each task together with task-appropriate evaluation metrics. All benchmark applications have been published on Flower Hub\textsuperscript{\ref{fn:apps_page}} for public access and reproducible evaluation (Appendix~\ref{appdix:hub_page}).

\textbf{Federated training}. For medical image segmentation, we employ a 3D U-Net~\citep{wang20193d} trained with cross-entropy loss using the Adam optimizer. For financial fraud detection, we use a multi-layer perceptron (MLP) with engineered tabular features, incorporating weighted sampling and loss weighting to address severe class imbalance~\citep{mishra2014comparative}. For legal instruction tuning, we adopt SmolLM3-3B as the base model and perform parameter-efficient fine-tuning using quantized LoRA, where only adapter weights are communicated between the server and clients~\citep{yue2023disc}. In these cross-silo settings, all five clients participate in each training round; we run 10 rounds for the legal instruction tuning task and 20 rounds for the medical and financial tasks.

For cross-device settings, phishing URL detection is implemented using a one-dimensional CNN trained with the AdamW optimizer~\citep{le2018urlnet}, with 10 out of 100 clients sampled per round over 20 rounds. For on-device audio tagging, raw audio signals are converted into log-mel spectrograms and processed using a compact CNN classifier~\citep{choi2016automatic}; 15 out of 50 clients are sampled per round, and training is conducted for 100 rounds. Across all tasks, we employ a cosine annealing learning-rate schedule. Detailed hyperparameter configurations are provided in the Appendix~\ref{appdix:hyper_params}.

\textbf{Evaluation metrics}. For medical image segmentation, we report the Dice score as the primary evaluation metric. For financial fraud detection, we use PR-AUC to account for extreme class imbalance. For legal instruction tuning, performance is evaluated using token-level F1 score. For phishing URL detection, we report ROC-AUC, and for audio tagging, we use classification accuracy.

\textbf{Federated optimization strategies}. We benchmark six widely used aggregation methods across all tasks. \textbf{FedAvg}~\citep{mcmahan2017communication} serves as the baseline, while \textbf{FedProx}~\citep{li2020federated} mitigates client drift under non-IID data. \textbf{FedAvgM}~\citep{hsu2019measuring} introduces server-side momentum for improved convergence. The FedOpt family, \textbf{FedAdam}, \textbf{FedAdagrad}, and \textbf{FedYogi}~\citep{reddi2020adaptive}, applies adaptive optimization on the server side with coordinate-wise learning rates and improved stability mechanisms. 
For fair comparison, we use a shared set of hyperparameters and avoid extensive task-specific tuning.

\textbf{System performance evaluation}. In addition to model performance, we record system-level metrics to characterize computational cost, communication overhead, memory usage, and runtime. For each client, we measure peak CPU and GPU memory consumption during local training to estimate hardware requirements.
We also track client training time per round, capturing the computational cost of model updates. To assess overall efficiency, we record the end-to-end wall-clock duration of each federated round and decompose it into training and non-training time, where the latter includes communication, synchronization, and orchestration overheads.
Finally, we measure the total communication volume exchanged between the server and clients, including model broadcasts and update aggregation. Together, these metrics provide a comprehensive view of the system-level characteristics of each benchmark.


\textbf{Deployment evaluation}. Deployment evaluation verifies that each benchmark can run beyond simulation using Flower’s deployment workflow. In this setting, clients are launched as independent SuperNodes with local data partitions, reflecting real-world federated environments. We conduct a full deployment run using the prescribed training rounds and client participation settings, exercising the complete distributed workflow, including server and client applications, data loading, local training, aggregation, communication, and metric reporting. A run is considered successful if the configured training procedure completes and participating clients return valid updates. Where possible, we compare deployment results with simulation under the same configuration to ensure consistency in partitioning, initialization, and metric reporting. Overall, this evaluation validates both reproducibility in simulation and executability under realistic deployment conditions.

\section{Experimental Evaluation of Benchmarks}
We evaluate each benchmark along three dimensions: model performance, system efficiency, and deployment behavior. These respectively assess predictive quality, computational and communication costs, and the ability to run reliably beyond simulation using independently deployed Flower SuperNodes without code modifications.

\begin{table}[t]
    \caption{\small Comparison of aggregation strategies across five tasks. The evaluation metrics are Dice score, PR-AUC, token-level F1 score, ROC-AUC, and accuracy for medical image segmentation, financial fraud detection, legal instruction tuning, phishing URL detection, and audio tagging, respectively. All experiments are conducted over three runs with different random seeds, reporting the mean performance with standard deviation in parentheses. The best results are highlighted in \textbf{bold}, and the second-best are \underline{underlined}. }
    \label{tab:main_results}
    \centering
    \resizebox{0.8\columnwidth}{!}{
    \begin{tabular}{lccccc}
         \toprule
          & Medical (\%) & Finance (\%) & Legal (\%) & URL (\%) & Audio (\%) \\
          \hline
         FedAvg & \underline{78.57 (0.09)} & \textbf{56.40 (0.45)} & 72.15 (0.63) & \underline{98.45 (0.05)} & \underline{48.70 (1.03)}  \\
         FedProx & \textbf{78.58 (0.28)} & \underline{55.21 (1.39)} & \underline{72.29 (1.44)} & \textbf{98.48 (0.01)} & \textbf{52.56 (3.71)}  \\
         FedAvgM & 78.31 (0.39) & 54.86 (0.56) & 71.96 (1.19) & 98.38 (0.02) & 47.44 (2.15)  \\
         FedAdam & 73.57 (3.76) & 14.71 (17.02) & 70.02 (1.85) & 93.64 (0.20) & 25.78 (5.68)  \\
         FedAdagrad & 64.02 (10.29) & 44.80 (2.37) & \textbf{72.97 (1.50)} & 91.02 (0.47) & 22.54 (2.66)  \\
         FedYogi & 75.18 (1.85) & 2.88 (0.10) & 70.19 (1.41) & 94.80 (0.18) & 27.80 (5.70)  \\
         
         \bottomrule
    \end{tabular}
    }
\end{table}

\begin{figure}[t]
\centering
    \includegraphics[width=\linewidth]{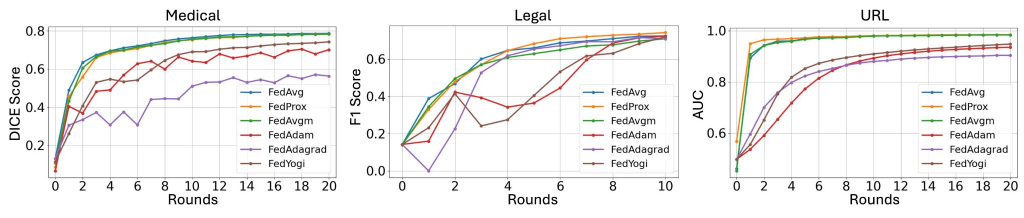}
\caption{\small Changes in evaluation metrics over FL rounds for the medical image segmentation, legal instruction tuning, and phishing URL detection tasks.}
\label{fig:model_line}
\end{figure}

\subsection{Model Performance Evaluation}

Our primary objective is to demonstrate Flower Hub as a benchmarking platform rather than to exhaustively optimize model performance. Accordingly, we report baseline results without extensive hyperparameter tuning; all benchmarks are publicly available for further community exploration. Table~\ref{tab:main_results} summarizes the evaluation results across the five benchmark tasks.

Across aggregation strategies, no single method consistently dominates, reflecting the challenges of non-IID client data. FedProx achieves the best performance in most cases, with FedAvg and FedAvgM performing similarly and showing comparable convergence behavior (Figure~\ref{fig:model_line}). In contrast, the FedOpt family (FedAdam, FedAdagrad, and FedYogi) often exhibits suboptimal or unstable performance, except in the legal instruction tuning task, where strong pretrained initialization improves results. This suggests that FedOpt methods are sensitive to hyperparameters (e.g., server learning rate and momentum), particularly when training from scratch~\citep{reddi2020adaptive,wu2023faster}. The issue is further exacerbated in tasks like financial fraud detection, where extreme class imbalance increases optimization difficulty (Figure~\ref{fig:label_dist}).\looseness-1

Performance also varies across tasks. In cross-silo settings, medical image segmentation and legal instruction tuning achieve relatively strong results (generally above 70\%), benefiting from full client participation and sufficient aggregate data. In contrast, financial fraud detection performs worse due to severe non-IID characteristics. In cross-device settings, phishing URL detection achieves high performance (above 90\% across methods), reflecting the relative simplicity of the task. Audio tagging, however, remains challenging, with all methods achieving only modest accuracy (slightly above 50\%), due to both heterogeneous client distributions and the intrinsic difficulty of audio representation learning.\looseness-1

Overall, these results establish baseline performance on the proposed benchmarks, providing a reference point for future research. They also illustrate the ease of conducting federated benchmarking experiments using Flower Hub.




\subsection{System Performance Evaluation}

Flower Hub includes a built-in system monitor for real-time performance tracking during benchmark execution. Table~\ref{tab:sys_results} summarizes system metrics across tasks in simulation mode using FedAvg. Differences across aggregation strategies are negligible, and thus omitted.

Communication cost is primarily determined by the size of the trainable model parameters. For example, the Medical task incurs the highest communication overhead due to the large 3D U-Net model, whereas the Finance task requires minimal communication (on the order of 0.01 GB) owing to its lightweight MLP architecture. CPU and GPU memory utilization reflect both data loading and model training processes, suggesting potential opportunities for optimization through improved hardware utilization and data pipeline design.

Client-side training time is influenced by both dataset size and model complexity. As expected, the legal tuning task, based on LLM fine-tuning, exhibits the longest training time, while the URL task requires comparatively little computation. Notably, training time does not dominate the total end-to-end latency across tasks. Instead, data preprocessing and communication overhead contribute significantly to overall execution time in federated settings.
These observations highlight the importance of jointly considering model design, data handling, and system-level efficiency when evaluating federated learning workloads.

\begin{table}[t]
    \caption{\small System performance across benchmark tasks in simulation mode, measured using FedAvg. The legal task is executed on an H200 GPU, while all other tasks use an A100 GPU. “Comm.” denotes the total communication cost over all training rounds. “CPU Mem.” and “GPU Mem.” represent the average peak memory usage across rounds. “Training” indicates the cumulative client training time, and “Latency” refers to the total end-to-end time required to complete federated training. }
    \label{tab:sys_results}
    \centering
    \resizebox{0.75\columnwidth}{!}{
    \begin{tabular}{lccccc}
         \toprule
          Tasks & Comm. (GB) & CPU Mem. (GB) & GPU Mem. (GB) & Training (s) & Latency (s) \\
          \hline
         Medical & 3.85 & 36.23 & 2.45 & 10.05 K & 40.52 K \\
         Finance & 0.01 & 2.78 & 0.02 & 363.01 & 13.25 K \\
         Legal & 2.86 & 9.47 & 21.24 & 14.63 K & 42.43 K  \\
         URL & 0.85 & 1.91 & 0.31 & 27.34 & 244.00   \\
         Audio & 0.28 & 7.33 & 0.10 & 156.67 & 2.17 K   \\
         
         \bottomrule
    \end{tabular}
    }
    \vspace{-2mm}
\end{table}

\subsection{Deployment Evaluation}

\begin{figure*}[t]
\centering
    \includegraphics[width=0.8\textwidth]{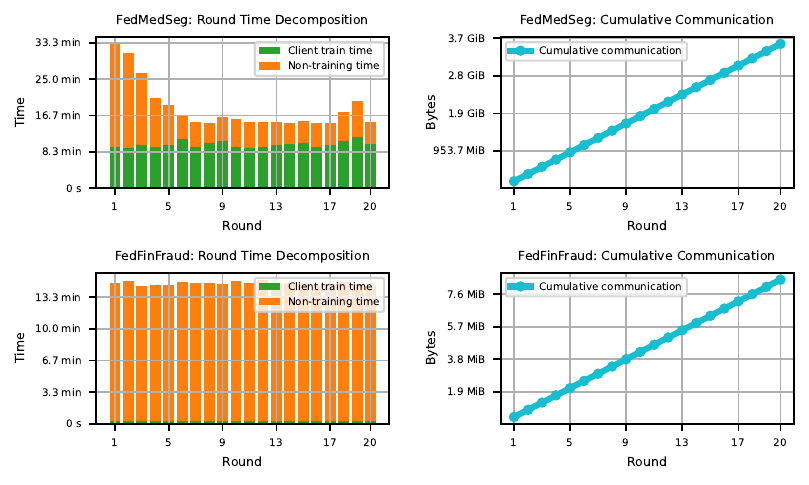}
\caption{\small Deployment profiles for FedMedSeg and FedFinFraud under FedAvg. Stacked bars show per-round wall-clock time split into client training and non-training overhead (e.g., preprocessing, communication, and synchronization). Right panels show cumulative communication over FL rounds.}
\label{fig:deployment_profiles}
\vspace{-2mm}
\end{figure*}

All benchmark applications on Flower Hub can be executed in a deployment setting without requiring code modifications. To validate this capability, we construct a cross-region deployment topology (Figure~\ref{fig:deployment_topology} \& Table~\ref{tab:deployment_instances}). 
Figure~\ref{fig:deployment_profiles} presents the evolution of wall-clock time and cumulative communication cost over federated learning rounds. The results illustrate the relative contribution of client-side training time within each round, as well as the overall system behavior during distributed execution. In particular, communication cost increases approximately linearly with the number of rounds, reflecting the use of a fixed model architecture and consistent parameter exchange across iterations.
These findings demonstrate the practical deployability of Flower Hub benchmarks and provide insight into the temporal and communication characteristics of real-world federated training.

\section{Related Work}

Research on federated benchmarking can be broadly grouped into three strands: benchmark suites, frameworks and tooling, and evaluation or reproducibility studies.

\textbf{Benchmark suites}. Early work such as LEAF~\citep{caldas2018leaf} established a foundation of federated datasets, metrics, and reference implementations. Subsequent efforts have expanded realism and domain coverage, including OARF~\citep{hu2022oarf}, FLBench~\citep{liang2020flbench}, FedScale~\citep{lai2022fedscale}, FLamby~\citep{ogier2022flamby}, and pFL-Bench~\citep{chen2022pfl}. More recent benchmarks target emerging settings such as multimodal learning, IoT data, and federated large language models~\citep{feng2023fedmultimodal,alam2023fedaiot,ye2024fedllm}. While these works improve realism and diversity, they are typically released as standalone repositories with tightly coupled infrastructure, limiting reuse and standardization.\looseness-1

\textbf{Frameworks and tooling}. A large body of work focuses on simplifying the implementation and execution of federated learning, including TensorFlow Federated~\citep{tensorflow2019tff}, FedLab~\citep{zeng2023fedlab}, FLUTE~\citep{garcia2022flute}, FedML~\citep{he2020fedml}, EasyFL~\citep{zhuang2022easyfl}, and FederatedScope~\citep{xie2022federatedscope}. Systems such as OpenFL~\citep{reina2021openfl}, NVFlare~\citep{roth2022nvidia}, and APPFL~\citep{ryu2022appfl} further emphasize deployment and privacy-preserving workflows. These frameworks reduce engineering overhead but primarily serve as execution substrates rather than defining benchmarks as standardized, portable artifacts.

\textbf{Evaluation and reproducibility}. Prior work has also addressed evaluation methodology and reproducibility. FedEval~\citep{chai2020fedeval} proposed a comprehensive evaluation framework, while UniFed~\citep{liu2022unifed} introduced schema-based comparability across multiple FL systems. MedPerf~\citep{karargyris2023federated} moves closer to executable benchmarking with standardized packaging and real-world deployment in healthcare. Other studies~\citep{prigent2025reproducibility} highlight the gap between simulation and deployment. However, these efforts either focus on evaluation standardization or domain-specific solutions, leaving benchmark packaging and portability largely unaddressed.

Against this background, Flower Hub is best understood not as another benchmark suite or framework, but as a benchmark platform built on top of Flower~\citep{beutel2020flower}. It decouples application logic from infrastructure, packages benchmarks as executable and versioned units, and enables seamless execution across simulation and deployment. This complements existing work by addressing the missing systems layer for benchmark exchange, portability, and long-term reusability.

\section{Conclusion}

We presented Flower Hub, a reproducible benchmarking platform for federated and decentralized learning that packages benchmarks as executable, versioned applications with standardized metadata and evaluation workflows. By decoupling application logic from infrastructure, Flower Hub enables seamless execution across simulation and deployment without code changes.
We demonstrated this approach through five realistic benchmarks spanning cross-silo and cross-device settings across multiple domains. Results across six aggregation strategies show strong task-dependent performance variation, while built-in system monitoring provides visibility into communication, memory, runtime, and deployment behavior. These findings highlight Flower Hub’s potential as a foundation for reproducible and portable federated benchmarking.

\textbf{Limitations and future work}. Our study emphasizes baseline evaluation rather than extensive optimization. The current benchmarks do not yet cover all FL settings, such as personalization, privacy, robustness, or highly constrained environments, and deployment experiments remain limited in scale. Future work will expand the benchmark suite, incorporate broader evaluations, and support large-scale, long-running deployments.

{
\bibliographystyle{abbrvnat}
\bibliography{neurips}
}

\clearpage
\appendix

\section{Flower Infrastructure}
\label{appdix:flwr_infra}

\begin{figure}[h]
\centering
    \includegraphics[width=\linewidth]{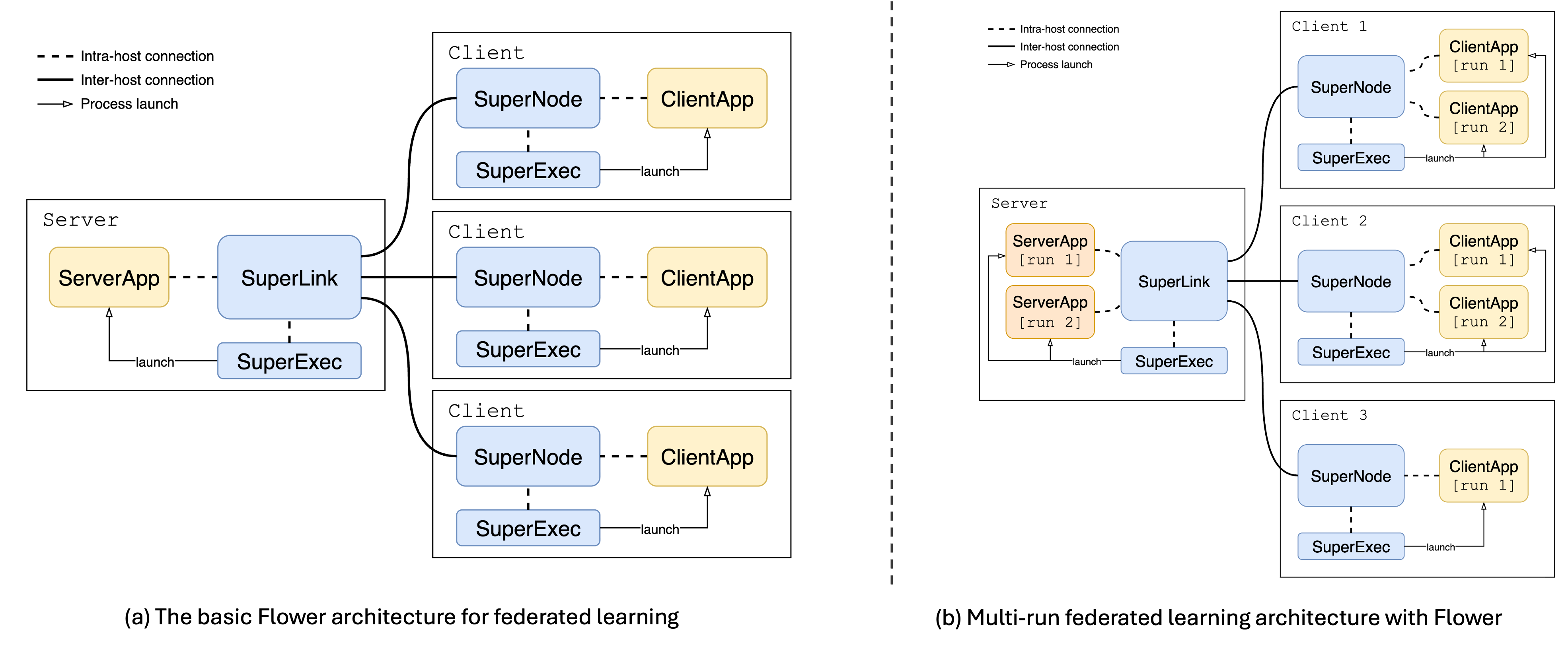}
\caption{\small Overview of Flower infrastructure.}
\label{fig:flwr_infra}
\end{figure}

Flower’s infrastructure separates long-lived system components responsible for networking and coordination from short-lived application components that implement project-specific federated learning logic. In a typical federated learning system, a central server coordinates training while multiple clients execute tasks on local data and return results. Flower adopts a hub-and-spoke architecture in which both server-side and client-side functionalities are divided into infrastructure processes and application processes. This separation enables multiple federated learning applications to share the same federation while using different models, hyperparameters, aggregation strategies, or machine learning frameworks (Figure~\ref{fig:flwr_infra}). Further details can be found in the Flower Documentation\footnote{\scriptsize\url{ https://flower.ai/docs/framework/index.html}\label{fn:flwr_doc}}.

\textbf{SuperLink}. On the server side, the core infrastructure component is the \textit{SuperLink}, a long-running process that acts as the communication hub of the federation. It forwards task instructions from the server-side application to connected client nodes and collects their results. Users interact with the SuperLink via the Flower command-line interface, while client-side SuperNodes connect to it through the Fleet API. The SuperLink is responsible for coordination and communication, rather than implementing learning logic.

\textbf{ServerApp}. The server-side learning logic is encapsulated in the \textit{ServerApp}, a short-lived process that defines the behavior of a federated learning run. This includes client selection, configuration, aggregation of client updates, and the implementation of federated optimization strategies (e.g., FedAvg). The ServerApp communicates with the SuperLink through the ServerAppIO API, allowing it to exchange messages with clients without managing low-level networking details.

\textbf{SuperNode}. On the client side, the corresponding infrastructure component is the \textit{SuperNode}, a long-running process that connects to the SuperLink, retrieves tasks, executes them, and returns results. Each SuperNode represents a participating client environment, such as a device, machine, or institutional data silo. Importantly, SuperNodes initiate outbound connections to the SuperLink and do not require inbound connectivity, making them suitable for deployment in restricted or firewall-protected environments.

\textbf{ClientApp}. Client-side computation is implemented in the \textit{ClientApp}, a short-lived process containing application-specific logic such as local training, evaluation, preprocessing, and access to local data. The ClientApp is executed on demand when a SuperNode is selected to participate in a round. It communicates with the SuperNode through the ClientAppIO API, while the SuperNode manages all network communication with the SuperLink. This separation allows the ClientApp to focus solely on local computation and data handling.

\textbf{SuperExec}. Flower also provides \textit{SuperExec}, a long-running process responsible for scheduling, launching, and managing application processes such as ServerApp and ClientApp. On the server side, SuperExec manages ServerApp instances in coordination with the SuperLink; on the client side, it manages ClientApp instances in coordination with the SuperNode. By default, SuperExec is launched automatically, although it can also be managed as a separate component if needed.

During a deployed run, the workflow proceeds as follows. A federation is first established by starting a SuperLink and connecting one or more SuperNodes. When a user invokes \texttt{flwr run}, the run is submitted through the SuperLink. The ServerApp is then launched to coordinate the training process, while selected SuperNodes execute the corresponding ClientApp locally. Communication between server and clients occurs indirectly via the SuperLink and SuperNodes, and results are returned to the ServerApp for aggregation.

\textbf{Simulation and Deployment}. Flower distinguishes between two execution modes: Simulation Runtime and Deployment Runtime. In the Simulation Runtime, \texttt{flwr run} executes the workflow locally using a managed SuperLink and multiple simulated clients, typically as worker processes on a single machine. This mode supports rapid prototyping, debugging, and algorithm validation. In the Deployment Runtime, SuperLink and SuperNodes operate as independent processes across distributed environments, communicate via secure (TLS-enabled) gRPC, and interact with real client-side data stored on devices or local systems. Crucially, the same ServerApp and ClientApp code can be used in both modes, with the runtime selected through configuration.

Overall, Flower’s infrastructure follows a layered design: SuperLink and SuperNodes provide persistent communication and coordination, SuperExec manages application lifecycles, and ServerApp and ClientApp implement task-specific learning logic. This separation decouples reusable infrastructure from application code, enabling multiple federated learning applications—including those distributed via Flower Hub—to share the same federation while flexibly selecting different subsets of clients for each run.

\begin{figure}[t]
\centering
    \includegraphics[width=\linewidth]{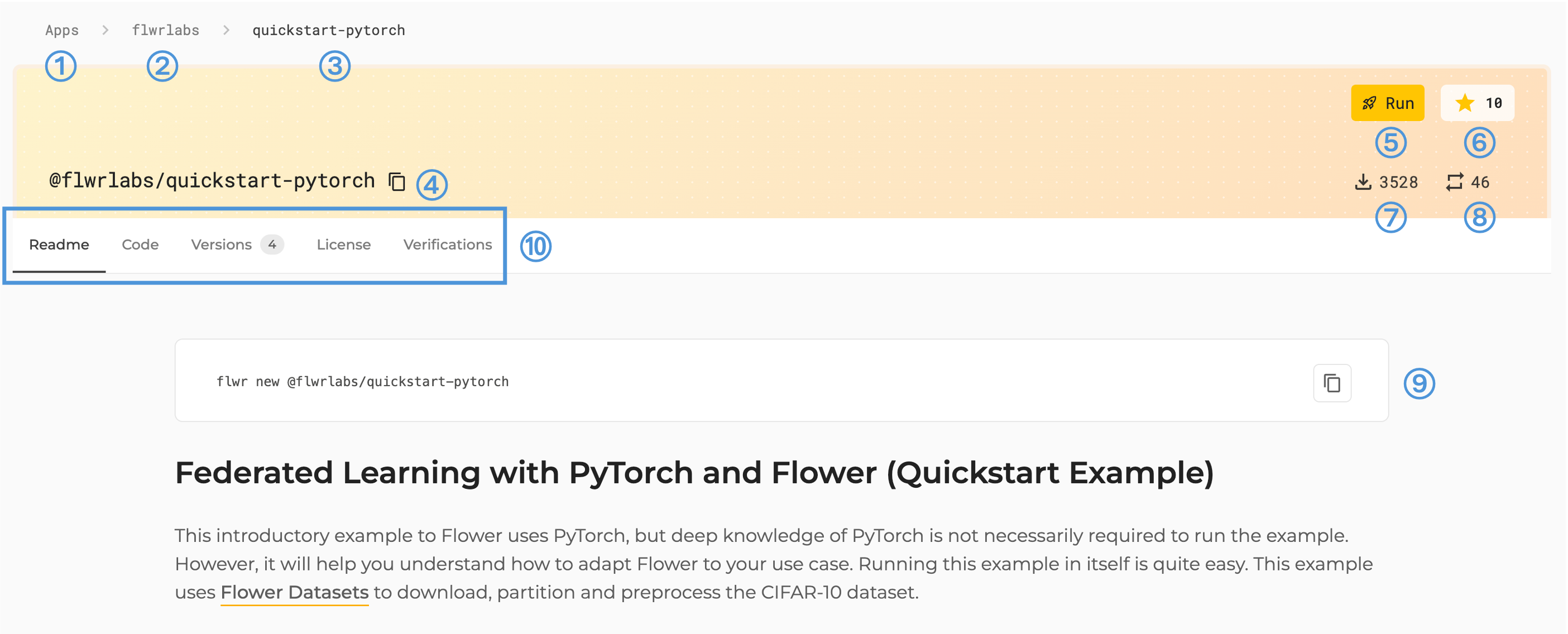}
\caption{\small Example of an application detail page on Flower Hub. \textcircled{1} indicates the navigation link to the main applications page. \textcircled{2} denotes the publisher account name, which links to the publisher’s profile page. \textcircled{3} displays the application name, and \textcircled{4} shows the full application identifier in the \texttt{@account/app} format. \textcircled{5} provides a button to execute the application on SuperGrid. \textcircled{6} reports the number of stars received by the application, while \textcircled{7} and \textcircled{8} indicate the total number of downloads and executions, respectively. \textcircled{9} presents the command for fetching the application to a local environment. \textcircled{10} corresponds to the information panel, including the README, code details, versioning, license, and verification metadata. The interface shown reflects the design at the time of publication and may evolve in future versions.}
\label{fig:app_page}
\end{figure}

\section{Flower Datasets}
\label{appdix:flwr_datasets}

Flower Datasets, provided through the \texttt{flwr-datasets} library, supports data preparation for federated learning, federated analytics, and federated evaluation. It enables centralized datasets to be transformed into client-specific partitions, facilitating the simulation of federated environments with either independent and identically distributed (IID) or heterogeneous non-IID data distributions. Further details are available in the Flower Datasets Documentation\footnote{\scriptsize\url{https://flower.ai/docs/datasets/}\label{fn:flwr_datasets_doc}}.

The core abstraction is the \texttt{FederatedDataset}, which integrates dataset loading, preprocessing, and partitioning. It supports direct loading from the Hugging Face Datasets Hub and partitions selected dataset splits into multiple client subsets using configurable partitioning strategies. For instance, \texttt{IidPartitioner} enables IID partitioning, while \texttt{DirichletPartitioner} and \texttt{PathologicalPartitioner} are commonly used to simulate non-IID client distributions.

Flower Datasets also supports centralized evaluation by allowing the training split to be partitioned while keeping the test split unpartitioned. The resulting partitions are represented as Hugging Face \texttt{Dataset} objects, ensuring compatibility with widely used machine learning frameworks such as PyTorch, TensorFlow, NumPy, Pandas, and JAX.

Beyond public datasets, Flower Datasets can operate on local data sources, including CSV, JSON, image, audio, and in-memory formats. For deployment-oriented scenarios, the \texttt{flwr-datasets create} command can generate pre-partitioned datasets on disk, enabling individual SuperNodes to access their assigned local data partitions efficiently.

Overall, Flower Datasets provides a reproducible and flexible data preparation layer for Flower-based experiments, bridging public and local datasets with configurable partitioning strategies to support both controlled benchmarking and realistic federated learning scenarios.

\begin{figure}[t]
\centering
    \includegraphics[width=\linewidth]{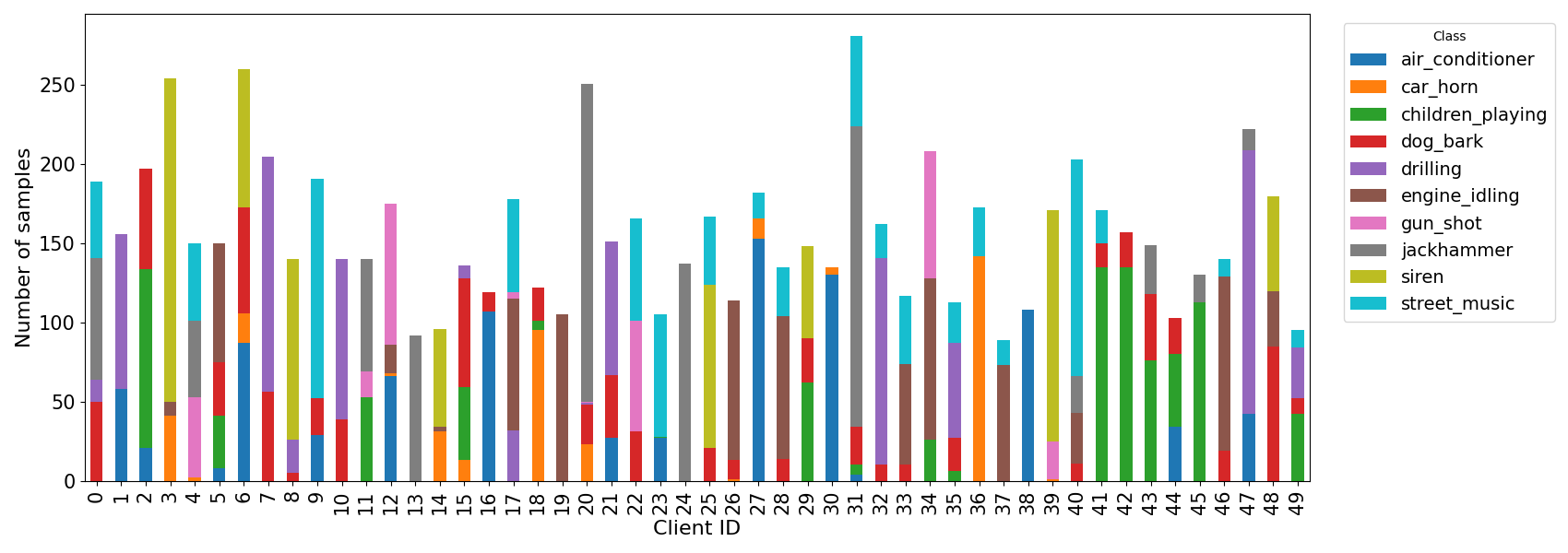}
\caption{\small Number of samples and class distribution across clients in the proposed federated audio tagging dataset.}
\label{fig:audio_by_class}
\end{figure}

\begin{figure}[t]
\centering
    \includegraphics[width=\linewidth]{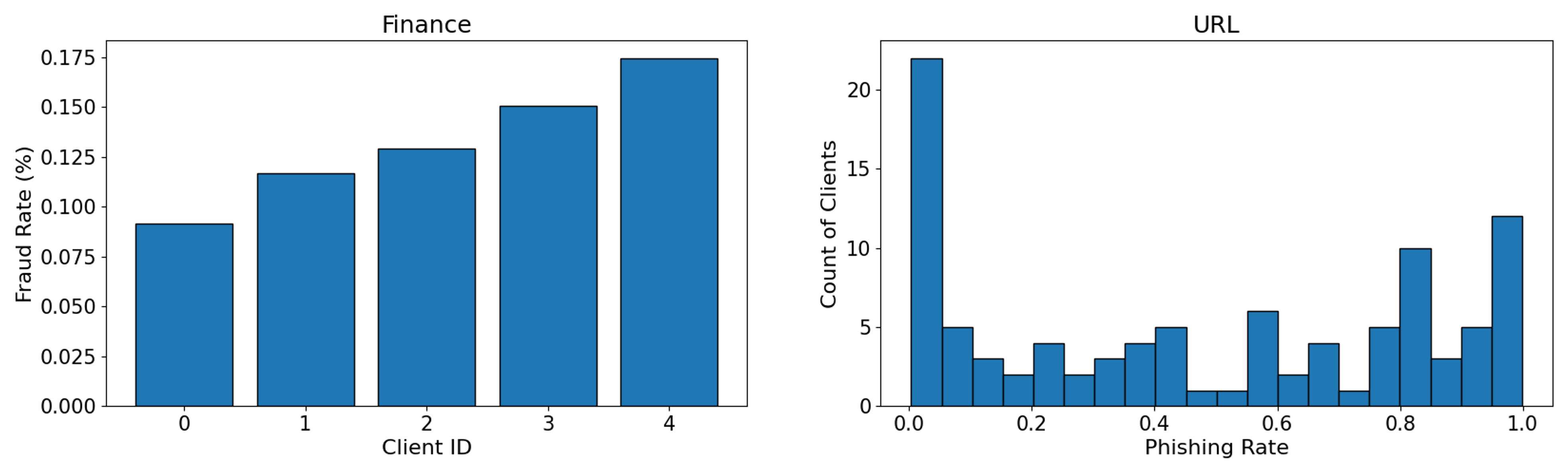}
\caption{\small Distribution of fraud rates and phishing rates across clients for the financial fraud detection and phishing URL detection benchmarks.}
\label{fig:label_dist}
\end{figure}

\section{How-to Guide for Flower Hub}
\label{appdix:hub_howto}

This section provides a practical guide to interacting with Flower Hub, including how to fetch, run, and publish benchmark applications. For additional details, we refer readers to the Flower Hub Documentation\footnote{\scriptsize\url{ https://flower.ai/docs/hub/index.html}\label{fn:hub_doc}}.

\subsection{Fetching an Application from Flower Hub}

To obtain an application from Flower Hub, users must first install the Flower command-line interface, \texttt{flwr}, within their local Python environment. Once installed, an application can be fetched using the command \texttt{flwr new @account/app}. For example, \texttt{flwr new @flwrlabs/quickstart-pytorch} downloads the specified application and initializes it locally.

Applications on Flower Hub may include metadata specifying the required Flower App Bundle (FAB) format version and compatible Flower versions. As a result, the fetching process may fail prior to download if the local Flower installation is incompatible with the selected application. Users are therefore advised to verify version compatibility before fetching.

\subsection{Running an Application from Flower Hub}

Applications hosted on Flower Hub can be executed in either the Simulation Runtime or the Deployment Runtime without modifying the application source code, using the command \texttt{flwr run @account/app}.

For deployment, users must configure and launch one or more SuperNodes. Each SuperNode requires the address of the corresponding SuperLink and, when necessary, a path to local datasets specified via the \texttt{--node-config} option. Configuration details may vary across applications; therefore, users should consult the application’s README for task-specific parameters and setup instructions.

\begin{table}[t]
    \caption{\small Selected hyperparameter configurations for the five benchmark tasks. }
    \label{tab:hyper_params}
    \centering
    \resizebox{\columnwidth}{!}{
    \begin{tabular}{lccccccc}
         \toprule
          Tasks & num-rounds & fraction-train & local-epochs & batch-size & lr-max & lr-min & optimizer \\
          \hline
         Medical & 20 & 1.0 & 3 & 4 & 1e-4 & 1e-5 & Adam \\
         Finance & 20 & 1.0 & 2 & 1024 & 1e-4 & 1e-5 & Adam \\
         Legal & 10 & 1.0 & 1 & 16 & 5e-5 & 1e-5 & AdamW  \\
         URL & 20 & 0.1 & 1 & 128 & 1e-3 & 1e-4 & AdamW   \\
         Audio & 100 & 0.3 & 2 & 32 & 1e-2 & 1e-4 & Adam  \\
         
         \bottomrule
    \end{tabular}
    }
\end{table}

\subsection{Publishing an Application on Flower Hub}

To publish an application on Flower Hub, it must be structured as a complete Flower project, including source code, documentation, and metadata. Applications should be designed to run in both Simulation and Deployment runtimes without modification to core logic. This is typically achieved by separating runtime-specific data-loading components from shared training logic. The accompanying README should document required configurations for both execution modes, including simulation settings and deployment data requirements.

Before publication, metadata must be defined in the \texttt{pyproject.toml} file. The \texttt{[project]} section should specify the application name, version, description, and license, while the \texttt{[tool.flwr.app]} section defines the publisher name and, if applicable, the FAB format and compatible Flower versions. The publisher field must match the Flower account used for publication. Notably, the application name is publicly visible and cannot be changed after initial release.

Flower Hub publishes source files rather than prebuilt artifacts. When executing \texttt{flwr app publish}, Flower recursively collects files from the project directory, applies filtering rules (e.g., file types and \texttt{.gitignore}), and validates the file set prior to upload. Validation includes checks on encoding, file count, individual file size, and total package size. If validation fails, the process terminates before upload.

To publish, users must first authenticate via \texttt{flwr login supergrid}, which initiates a browser-based login linked to their Flower account. The application can then be published using \texttt{flwr app publish <your-app-path>}. Upon successful publication, the application becomes accessible on Flower Hub at a URL of the form \texttt{https://flower.ai/apps/<account>/<app>/}.

Subsequent updates require modifying the source code, incrementing the \texttt{version} field in \texttt{pyproject.toml}, and re-running the publication command. Flower Hub maintains all published versions, enabling users to access and execute specific releases as needed.

\subsection{Signing an Application from Flower Hub}

Flower Hub supports application signing as a mechanism for attaching reviewer-generated verification metadata to published applications. An application must first be available on Flower Hub before it can be signed; however, the reviewer need not be the original publisher. This enables independent users or organizations to review and verify applications published by others. We note that application signing and related verification metadata are currently provided as preview features and may evolve over time.

The signing process is based on a cryptographic public--private key pair. The reviewer generates a key pair, securely stores the private key, and registers the corresponding public key in their Flower account profile. After authentication, the reviewer can invoke the Flower CLI command \texttt{flwr app review} to inspect and sign an application. Signing can target either the latest version (e.g., \texttt{@account/app}) or a specific version (e.g., \texttt{@account/app==x.y.z}).

During the review process, the Flower CLI downloads the FAB, unpacks it for inspection, and prompts the reviewer to confirm the signing operation. The reviewer then provides the path to their private key, and a cryptographic signature is generated over the FAB digest together with a timestamp. This signature is submitted to Flower Hub along with the application identifier and version, ensuring that verification is tied to a specific application artifact.

A key property of this mechanism is its decentralized trust model. Flower Hub does not restrict verification to application publishers or a central authority. Instead, any authenticated user with a registered signing key can review and sign an application, including those published by other accounts. The resulting signatures are displayed on the application page, allowing users to inspect which entities have verified the application and to determine which signers they trust.

This decentralized verification approach is particularly relevant for federated learning, where applications are executed across multiple organizations, devices, or data-owning sites. By enabling independent reviewers to sign the same application version, Flower Hub supports a layered trust model in which institutions can rely on signatures from trusted parties. The verification section thus serves as a transparency mechanism: while it does not guarantee trustworthiness, it provides verifiable evidence that specific reviewers have inspected and cryptographically endorsed a given application version.

\begin{figure}[t]
\centering
    \includegraphics[width=\linewidth]{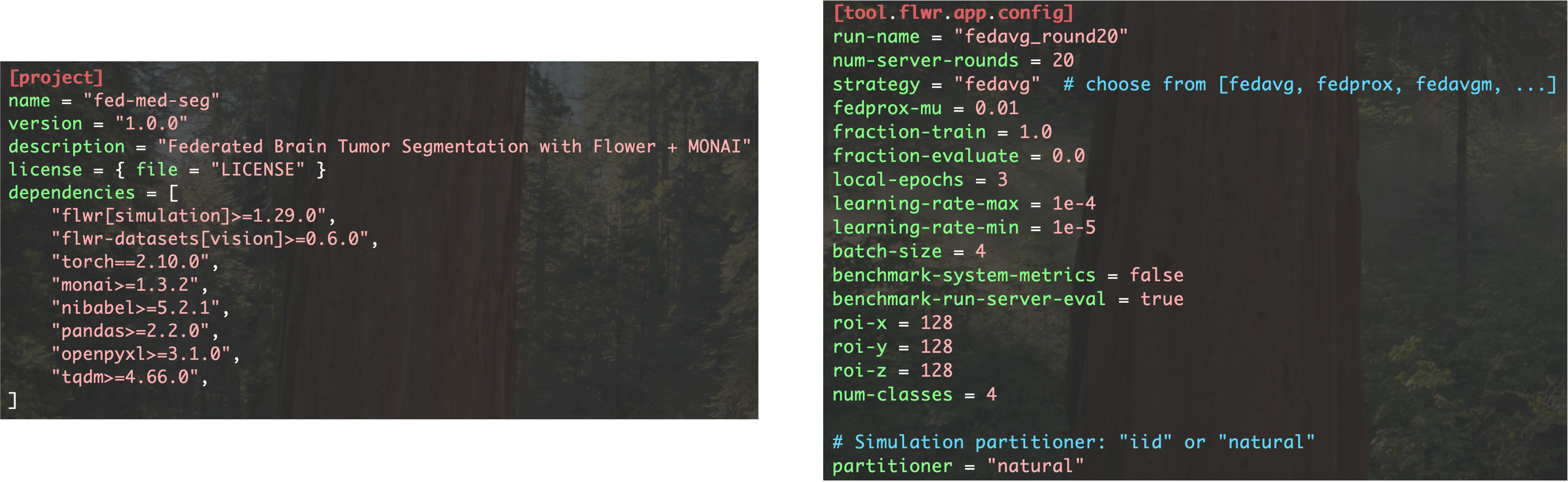}
\caption{\small Configuration example for the \texttt{fed-med-seg} application. (Left) The configuration block specifies metadata, including the application name, version, description, license, and pinned dependencies. (Right) The configuration block defines the hyperparameters used for federated training and evaluation.}
\label{fig:config_example}
\end{figure}

\section{Configuration Example for Flower Hub Applications}
\label{appdix:config_example}

Each benchmark application on Flower Hub follows a standardized package structure with a well-defined schema, explicit configurations, pinned dependencies, and machine-readable metadata. Figure~\ref{fig:config_example} illustrates an example configuration for the medical image segmentation application.

The configuration consists of two main blocks. The first defines application metadata, including the version, license, and pinned dependencies, which ensure reproducibility across environments. The second specifies the hyperparameters used for federated training and evaluation, providing flexibility to adapt experimental settings while maintaining a consistent structure.

\section{Application Detail Page on Flower Hub}
\label{appdix:hub_page}

All benchmark applications used in this paper are publicly available on Flower Hub. The corresponding application identifiers are as follows:

\begin{itemize}
    \item \texttt{@flwrlabs/fed-med-seg}\footnote{\scriptsize\url{https://flower.ai/apps/flwrlabs/fed-med-seg/}}
    \item \texttt{@flwrlabs/fed-fin-fraud}\footnote{\scriptsize\url{https://flower.ai/apps/flwrlabs/fed-fin-fraud/}}
    \item \texttt{@flwrlabs/fed-legal-llm}\footnote{\scriptsize\url{https://flower.ai/apps/flwrlabs/fed-legal-llm/}}
    \item \texttt{@flwrlabs/fed-phish-guard}\footnote{\scriptsize\url{https://flower.ai/apps/flwrlabs/fed-phish-guard/}}
    \item \texttt{@flwrlabs/fed-audio-tagging}\footnote{\scriptsize\url{https://flower.ai/apps/flwrlabs/fed-audio-tagging/}}
\end{itemize}

Each application on Flower Hub is associated with a dedicated detail page that provides comprehensive information, including metadata, source code, documentation, and usage statistics (Figure~\ref{fig:app_page}). This design ensures transparency and accessibility, making all benchmark applications fully open source and facilitating community engagement, reuse, and extension. Such openness contributes to the robustness and sustainability of the Flower Hub ecosystem.

\begin{figure}[t]
\centering
    \includegraphics[width=\linewidth]{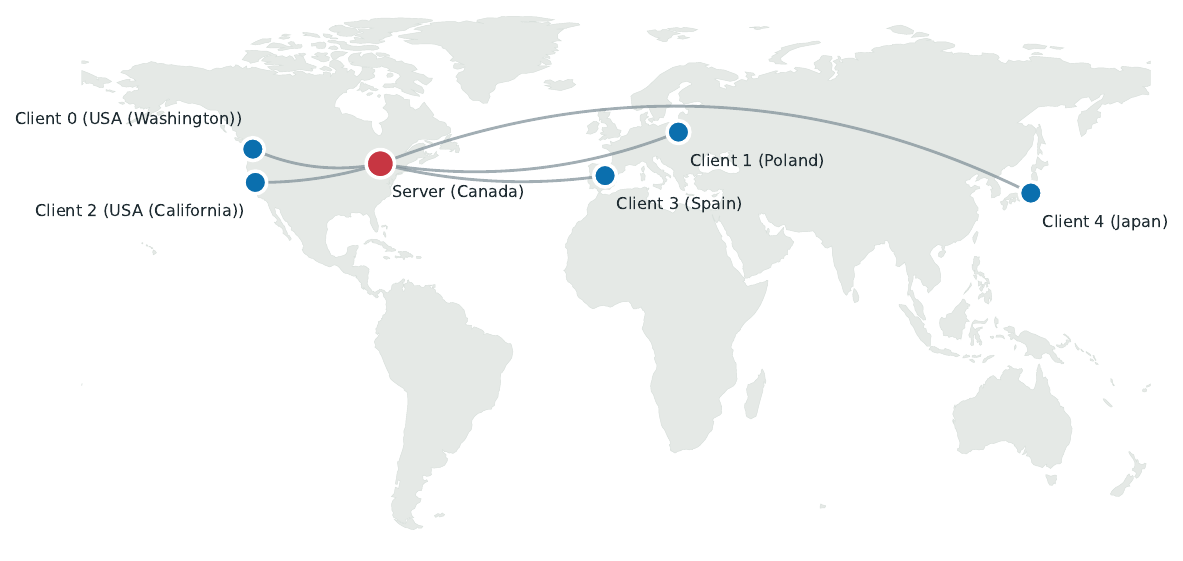}
\caption{\small Cross-region deployment topology used in our deployment-mode experiments. The federated server was hosted in Canada, while five clients were deployed on Vast.ai instances in the USA, Poland, Spain, and Japan. Edges indicate client-server communication during federated training.}
\label{fig:deployment_topology}
\end{figure}

\begin{table}[t]
\centering
\caption{\small Vast.ai instance configuration used for deployment-mode experiments.}
\label{tab:deployment_instances}
\scriptsize
\setlength{\tabcolsep}{3pt}
\resizebox{0.6\columnwidth}{!}{%
\begin{tabular}{llll}
\toprule
Role & GPU & CPU & VRAM \\
\midrule
Server   & 1$\times$ RTX 5090 & AMD Ryzen 9 5900XT 16-Core & 32 GB \\
Client 0 & 1$\times$ RTX 5090 & AMD EPYC 9J14 96-Core & 32 GB \\
Client 1 & 1$\times$ RTX 5090 & AMD Ryzen 7 7700X 8-Core & 32 GB \\
Client 2 & 1$\times$ RTX 5090 & AMD EPYC 9654 96-Core & 32 GB \\
Client 3 & 1$\times$ RTX 5090 & AMD EPYC 7642 48-Core & 32 GB \\
Client 4 & 1$\times$ RTX 4090 & AMD EPYC 7662 64-Core & 48 GB \\
\bottomrule
\end{tabular}%
}
\end{table}

\section{Detailed Benchmark Settings}

\subsection{Client-level Label Distribution of FL Datasets}
\label{appdix:label_dist}

Figures~\ref{fig:audio_by_class} and~\ref{fig:label_dist} illustrate the class distributions across clients for the audio tagging, financial fraud detection, and phishing URL detection benchmarks. These results highlight the pronounced non-IID characteristics of the proposed federated datasets.

For the audio tagging dataset, no single client contains the full set of classes, increasing the difficulty of learning a globally consistent model. In the financial fraud detection dataset, the class distribution is extremely imbalanced, with all clients exhibiting fraud rates below 0.2\%, reflecting realistic real-world conditions. Similarly, in the phishing URL detection dataset, the phishing rate varies substantially across clients, introducing additional heterogeneity. Together, these properties create challenging yet representative settings for evaluating federated learning algorithms.

\begin{figure}[t]
\centering
    \includegraphics[width=\linewidth]{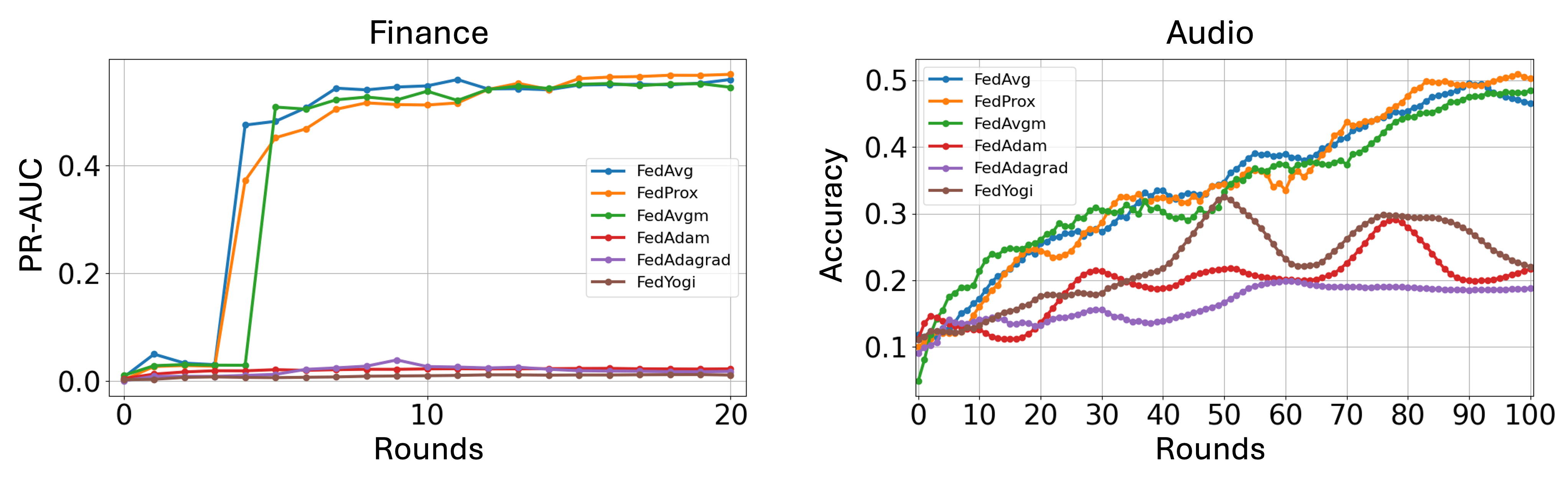}
\caption{\small Changes in evaluation metrics over federated learning rounds for the financial fraud detection and audio tagging tasks. FedAdam, FedAdagrad, and FedYogi exhibit suboptimal or unstable performance on these tasks.}
\label{fig:model_line2}
\end{figure}

\begin{table}[t]
\centering
\caption{\small Client-side memory footprint during deployment-mode FedAvg runs.}
\label{tab:deployment_memory}
\resizebox{0.6\columnwidth}{!}{%
\begin{tabular}{lrrrr}
\toprule
Benchmark & Mean CPU & Peak CPU & Mean GPU & Peak GPU \\
\midrule
FedMedSeg & 23.27 GiB & 23.31 GiB & 2.45 GiB & 2.45 GiB \\
FedFinFraud & 1.73 GiB & 1.74 GiB & 20.15 MiB & 20.15 MiB \\
\bottomrule
\end{tabular}%
}
\end{table}

\subsection{Hyperparameter Selection}
\label{appdix:hyper_params}

Table~\ref{tab:hyper_params} summarizes the common hyperparameter configurations used across the five benchmark tasks. The complete set of hyperparameters for each task is available on the corresponding application detail pages (Section~\ref{appdix:hub_page}). 

These configurations are chosen based on commonly adopted settings in prior literature, and we do not perform extensive hyperparameter tuning. The reported results should therefore be interpreted as baseline performance, providing a reference for future optimization and exploration.

\subsection{Deployment Configuration}

In deployment mode, we configure a set of heterogeneous GPU instances to approximate realistic cross-region federated learning environments. Figure~\ref{fig:deployment_topology} illustrates the deployment topology, in which the federated server is hosted in Canada, while five clients are distributed across the United States, Poland, Spain, and Japan.

Table~\ref{tab:deployment_instances} summarizes the hardware configurations used in this setup. The server is deployed on an RTX 5090 instance, while client nodes run on a mix of RTX 5090 and RTX 4090 instances. This configuration reflects a practical cross-silo FL scenario with geographically distributed participants and heterogeneous compute resources. We use this setup to validate that all benchmarks can be executed in a real deployment environment without requiring any modifications to the application code.

\section{Additional Experimental Results}

Figure~\ref{fig:model_line2} presents the evolution of evaluation metrics over federated learning rounds for the financial fraud detection and audio tagging tasks. FedAvg, FedProx, and FedAvgM exhibit similar convergence behavior, with FedProx achieving the best final performance on both tasks. In contrast, the FedOpt family (FedAdam, FedAdagrad, and FedYogi) performs poorly and exhibits instability. This can be attributed to two main factors: (1) both tasks involve highly non-IID client data distributions, which complicate optimization; and (2) FedOpt methods are sensitive to hyperparameter choices, requiring careful tuning to achieve stable and competitive performance.

For the deployment experiments, we additionally report both peak and average client-side memory usage in Table~\ref{tab:deployment_memory}. CPU memory consumption is primarily driven by data preprocessing and loading, whereas GPU memory usage is dominated by model training. The financial fraud detection task, which employs a lightweight MLP model, exhibits significantly lower GPU memory usage compared to the other tasks.



\end{document}